\documentclass[3p,times,procedia]{elsarticle}
\usepackage{amsmath}
\usepackage{booktabs}
\usepackage{ecrc}
\usepackage[bookmarks=false]{hyperref}
    \hypersetup{colorlinks,
      linkcolor=blue,
      citecolor=blue,
      urlcolor=blue}

\usepackage[T1]{fontenc}

\volume{00}

\firstpage{1}

\journalname{Procedia Computer Science}

\runauth{J. Bąba and J. A. Chudziak}

\jid{procs}

\usepackage{amssymb}

\usepackage[figuresright]{rotating}

\begin{document}
\begin{frontmatter}



\dochead{30th International Conference on Knowledge-Based and Intelligent Information \& Engineering Systems (KES 2026)}%

\title{From Argument Components to Graphs: A Multi-Agent Debate \\ with Confidence Gating for Argument Relations} 


\author{Jakub Bąba} 
\author{Jarosław A. Chudziak}

\address{Faculty of Electronics and Information Technology, Warsaw University of Technology, Poland}

\begin{abstract}
Large Language Models (LLMs) are increasingly assessed and utilized in the field of Argument Mining (AM), thanks to their strong general reasoning capabilities. However, standard training-free models often miss sophisticated details, specifically in contexts where two parts of the text have to be analyzed together. Furthermore, self-correction mechanisms tend to reinforce initial hallucinations in reasoning. Overcoming these limitations typically requires expensive, domain-specific supervised fine-tuning. Recent work has shown that a multi-agent paradigm can address such weaknesses for the component classification task through dialectical refinement with a Proponent-Opponent-Judge architecture, setting a promising direction for training-free approaches in the field. In this paper, we extend and evaluate this framework on the Argument Relation Identification and Classification (ARIC) task, reformulating it as a debate over component pairs. Besides that, we introduce a confidence gating mechanism that enables debating only on the uncertain cases and accepting the initial prediction when confidence is high. On the UKP Argument Annotated Essays v2 corpus, we demonstrate that the selective debate achieves the highest Macro F1 among all training-free methods, while debate over all samples degrades performance below that of one of the baselines. All generative approaches also outperform fine-tuned RoBERTa models on Macro F1, suggesting that the under-representation of the Attack class was more damaging to supervised fine-tuning than to inference-only models. Additionally, our framework produces human-readable debate transcripts, offering interpretability absent from both single-agent and supervised classifiers.

\end{abstract}

\begin{keyword}
Artificial Intelligence; Natural Language Processing; Argument Mining; Multi-Agent Systems; Large Language Models.




\end{keyword}
\cortext[cor1]{Corresponding author. Tel.: +0-000-000-0000 ; fax: +0-000-000-0000.}
\end{frontmatter}

\email{jakub.baba.stud@pw.edu.pl}



\section{Introduction}

\begin{figure}[t]
    \centering
    \includegraphics[width=\textwidth]{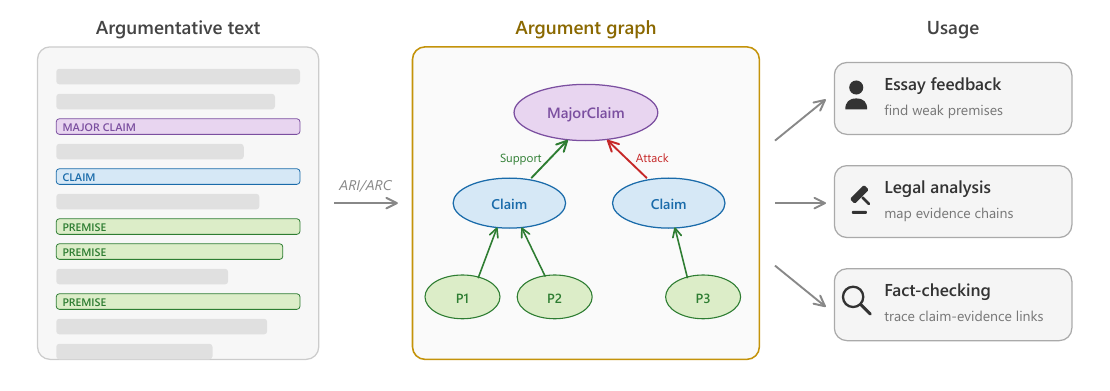}
    \caption{Overview of argument graph construction from annotated text and its downstream applications.}
    \label{fig:overview}
\end{figure}

Argument Mining - the automated extraction and identification of argumentative structures from text - is a crucial field for systems moving beyond surface-level text evaluation toward deep logical analysis~\cite{lawrence-reed-2019-argument}. A complete representation of argumentative discourse requires not only identifying singular arguments and their roles (e.g.\ Claim or Premise), but also mapping relations from one to another through directed Support and Attack links. This relational mapping, defined by Argument Relation Identification (ARI) and Classification (ARC), is the core of constructing argument graphs that comprehensively represent the details of the argumentative text~\cite{stab2017parsing}. The resulting graphs can support tasks such as automated essay scoring, legal analysis and claim verification, where systems can benefit from tracing particular argumentative paths in the argument tree (Figure~\ref{fig:overview}).

Large Language Models have become visible tools in Argument Mining, covering tasks from component detection to stance classification~\cite{li2025largelanguagemodelsargument}. Their strengths come from strong general reasoning and indispensable flexibility - they can be used without task-specific retraining. Recent work has demonstrated that appropriately prompted LLMs can approach supervised baselines on relation classification~\cite{gorur2025retrievalargumentationenhancedmultiagent}, while fine-tuned LLMs currently define the state of the art across AM subtasks~\cite{cabessa-etal-2025-argument}. In parallel, prior work has shown that a multi-agent debate paradigm — where a Proponent, Opponent, and Judge engage in structured dialectical refinement — can be successfully applied to Argument Component Classification, achieving results competitive with supervised approaches without any task-specific training~\cite{baba2026multiagentdialecticalrefinementenhanced}.

However, argumentation is relational at its core - while information about what individual components assert is valuable, the way the components connect is crucial. Relation classification demands context-sensitive reasoning about logical dependencies between two statements simultaneously, and errors in relation identification propagate into downstream graph construction. This setting is particularly prone to generative hallucination: LLMs may confidently assert logical links that do not genuinely exist in the text, a problem that single-agent self-correction mechanisms struggle to address due to sycophantic self-reinforcement~\cite{chen2025self}. It is not obvious whether a dialectical refinement paradigm can be transferred to this fundamentally harder relational setting, even though it has previously shown effective for component classification.

In this work, we investigate this question by extending the multi-agent debate framework from~\cite{baba2026multiagentdialecticalrefinementenhanced} to joint Argument Relation Identification and Classification (ARIC). The proposed approach formulates relation classification as a structured debate over component pairs, where a Judge adjudicates based on the debate transcript with explicit instructions to penalize hallucinated logical links. We further introduce a confidence gating mechanism that skips debate for high-certainty predictions, reducing computational cost. We evaluate the framework on the UKP Argument Annotated Essays v2 corpus~\cite{stab2017parsing} against single-agent inference baselines and fine-tuned RoBERTa models. Our results show that indiscriminate debate over all samples does not improve over the strongest baseline, but when combined with confidence gating, the dialectical paradigm achieves the highest Macro F1 among all of the methods. Additionally, all generative approaches outperform fine-tuned supervised models on Macro F1, due to the class imbalance being more damaging to them. Finally, the debate transcripts naturally provide a degree of interpretability absent from single-pass inference.

\section{Related Work}

Our work is situated at the intersection of three aspects of Argument Mining. We introduce Argument Relation Identification and Classification and review approaches to the argumentation tasks, both traditional and generative. At the end, we survey multi-agent reasoning systems and show their potential in the field.

\subsection{Argument Relation Identification and Classification}

Argument Mining contains several sequential subtasks, from boundary detection and component classification to the identification and labeling of relations~\cite{lawrence-reed-2019-argument}. While component-level tasks focus on categorizing isolated spans of text, constructing a complete argument graph requires structural mapping through Argument Relation Identification (ARI) - determining whether a link exists between two components; and Argument Relation Classification (ARC) - labeling that link as Support or Attack~\cite{stab2017parsing}. These relational subtasks are inherently harder than component classification, as they demand simultaneous reasoning about logical dependencies between pairs of statements within their broader discourse context.

Foundational approaches formulated ARI/ARC as pairwise classification, exploiting positional and syntactic features through SVMs~\cite{stab2017parsing,habernal-gurevych-2017-argumentation}. Neural architectures improved upon these baselines with structured SVMs combined with RNNs for global graph prediction~\cite{niculae-etal-2017-argument}, while more recent end-to-end formulations jointly address component extraction and relation classification through autoregressive structure prediction~\cite{das2025endtoendargumentminingautoregressive} or frame-semantic prompt templates~\cite{moslemnejad2025prompt}. Despite steady progress, relations remain sparsely distributed and context-dependent, and misclassified edges propagate through downstream graph construction — making robust relation inference a persistent challenge that motivates the exploration of alternative reasoning paradigms.

\subsection{Large Language Models in Argumentation}

Generative LLMs have opened the door to the exploration of the training-free alternatives in numerous fields, including argument mining~\cite{SADOWSKI20252166}. LLMs are used for component detection, stance classification and relation labeling through LLM-driven techniques alone, such as prompt-based reasoning. A recent comprehensive survey demonstrates that they have substantially advanced AM across such tasks, while identifying open long-term challenges, including model interpretability and long-context reasoning. A study of LLMs for relation based AM shows that appropriately prompted models can significantly outperform RoBERTa baselines~\cite{gorur2024largelanguagemodelsperform}, demonstrating that LLMs can handle nuanced relational judgments when carefully instructed. However, broader evaluations reveal that even state-of-the-art LLMs exhibit systematic failure modes, including sensitivity to prompt formulation, difficulty detecting implicit criticism, and misalignment between arguments and their target claims~\cite{pietron2026comprehensivestudyllmbasedargument}.

Recent work also includes supervised models as a response to the limitations of prompting alone. Fine-tuned large language models now achieve state-of-the-art results on standard AM corpora, outperforming earlier transformer baselines~\cite{cabessa-etal-2025-argument}, while instruction-tuning over broad sets of argumentation tasks enables zero-shot transfer to previously unseen problems~\cite{stahl2025arginstruct}. Multi-task training across diverse AM datasets has further demonstrated that joint optimization can preserve task-specific accuracy without catastrophic forgetting~\cite{savigny2025ameliafamilymultitaskendtoend}. However, these supervised strategies reintroduce a heavy dependence on high-quality annotated data, limiting their applicability in low-resource domains and languages where such annotations are rare. Moreover, fine-tuned models offer limited interpretability regarding individual predictions, making it difficult to trace why a particular argumentative relation was assigned.

\subsection{Multi-Agent Systems for Reasoning}

Multi-Agent Systems (MAS) leverage a collaborative-focused approach to enhance LLM problem-solving beyond what single-agent solutions can achieve~\cite{tran2025multiagentcollaborationmechanismssurvey}. MAS frameworks have shown particular promise across NLP and reasoning-intensive tasks~\cite{kostka-chudziak-2024-synergizing,gorur2025retrievalargumentationenhancedmultiagent}, benefiting from structured role assignment, shared state management, and the ability to decompose complex problems into specialized sub-tasks~\cite{estornell2025acccollabactorcriticapproachmultiagent}.

One of the well-documented limitations of single agent LLMs is their tendency towards sycophantic behavior: when asked to correct or verify their answers, models often reinforce initial errors rather than revising them~\cite{chen2025self}, a problem that sometimes persists even in multi-agent settings when agents don't challenge each other~\cite{pitre-etal-2025-consensagent}. Multi-Agent Debate addressed this by replacing self-reflection with inter-agent discussion. Multiple studies show that structured debate and critiquing each other's reasoning can improve accuracy~\cite{du2024improvingfactualityreasoninglanguage,liang-etal-2024-encouraging}. However, the effectiveness of the debate is not guaranteed. A NeurIPS 2025 spotlight paper~\cite{choi2025debatevoteyieldsbetter} disentangled MAD into Majority Voting and inter-agent Debate and found that most performance gains are the result of the voting alone. Their analysis showed that in order to make a meaningful contribution by debate alone, targeted architectural interventions are required.

Such interventions often take the form of structured role assignments and adversarial dynamics. Assigning diverse specialized roles to agents has been shown to outperform homogeneous configurations~\cite{Zhou2025AdaptiveHM}, and experiments confirm that intrinsic reasoning strength and group diversity are the dominant drivers of debate success, rather than structural configurations~\cite{wu2025llmagentsreallydebate}. These findings suggest that MAD architectures benefit the most when agents are explicitly forced into fixed positions and when final adjudication is handled by a dedicated evaluator rather than simple aggregation.

Debate frameworks are also successfully applied to argumentation. One framework was used to evaluate implicit premises, outperforming both neural baselines and single-agent LLMs~\cite{ku-etal-2025-multi}.Most directly relevant to our work MAD-ACC, a Proponent-Opponent-Judge framework, was utilized for the task of Argument Component Classification, outperforming all single-agent training-free approaches and narrowing the gap to supervised approaches~\cite{baba2026multiagentdialecticalrefinementenhanced}. The framework was especially effective at resolving Claim vs Premise ambiguity, where surface-level semantics were misleading for single-pass classifiers. Our work extends this architecture by moving from component-level to relation-level task where the reasoning has to consider pairs of components. This task introduces a risk of hallucinating relations that makes the adversarial approach especially valuable.

\section{Problem and Approach}

We present our adaptation of the multi-agent debate framework used for relation identification and classification. We reformulate the task as a structured debate over component pairs. Our adversarial approach is particularly applicable to this task: by showing both perspectives, the framework is designed to challenge hallucinated arguments through adversarial scrutiny.

\subsection{Task Formulation}

We formalize the joint task of Argument Relation Identification and Classification as a pair-wise labeling task over pre-identified argument components. Let $D = \{c_1, c_2, \ldots, c_n\}$ be an argumentative document consisting of $n$ argument components. For each ordered pair of components $(c_s, c_t)$, where $c_s$ denotes the source and $c_t$  the target, let $\mathcal{C}_{s,t}$ denote the shared context window (e.g. the enclosing paragraph or essay).

The objective is to establish a mapping function:

\[\Phi : (c_s, c_t, \mathcal{C}_{s,t}) \rightarrow y\]

that assigns the correct label $y \in \mathcal{Y}$, where:

\[\mathcal{Y} = \{\textit{Support}, \textit{Attack}, \textit{None}\}\]
The label set jointly addresses Argument Relation Identification and Classification:
\begin{itemize}
    \item \textbf{Support}: The source component $c_s$ provides evidence, reasoning, or justification that strengthens $c_t$.
    \item \textbf{Attack}: The source component $c_s$ undermines, contradicts, or challenges $c_t$.
    \item \textbf{None}: No directed argumentative relation exists between $c_s$ and $c_t$.
\end{itemize}

This formulation subsumes ARI and ARC into a single classification step: the \textit{None} label captures the identification decision (whether a link exists), while the \textit{Support} and \textit{Attack} labels capture the classification decision (what type of link it is). The directionality of the relation is encoded in the ordering of the pair — $\Phi(c_s, c_t, \mathcal{C}_{s,t})$ and $\Phi(c_t, c_s, \mathcal{C}_{s,t})$
are treated as independent predictions.

Following prior work on the UKP corpus~\cite{stab2017parsing,cabessa-etal-2025-argument}, candidate pairs are restricted to components within the same paragraph, reflecting the observation that the vast majority of argumentative relations in persuasive essays operate at the local discourse level.

\subsection{The Framework}
\begin{figure}[]
    \centering
    \includegraphics[width=0.8\textwidth]{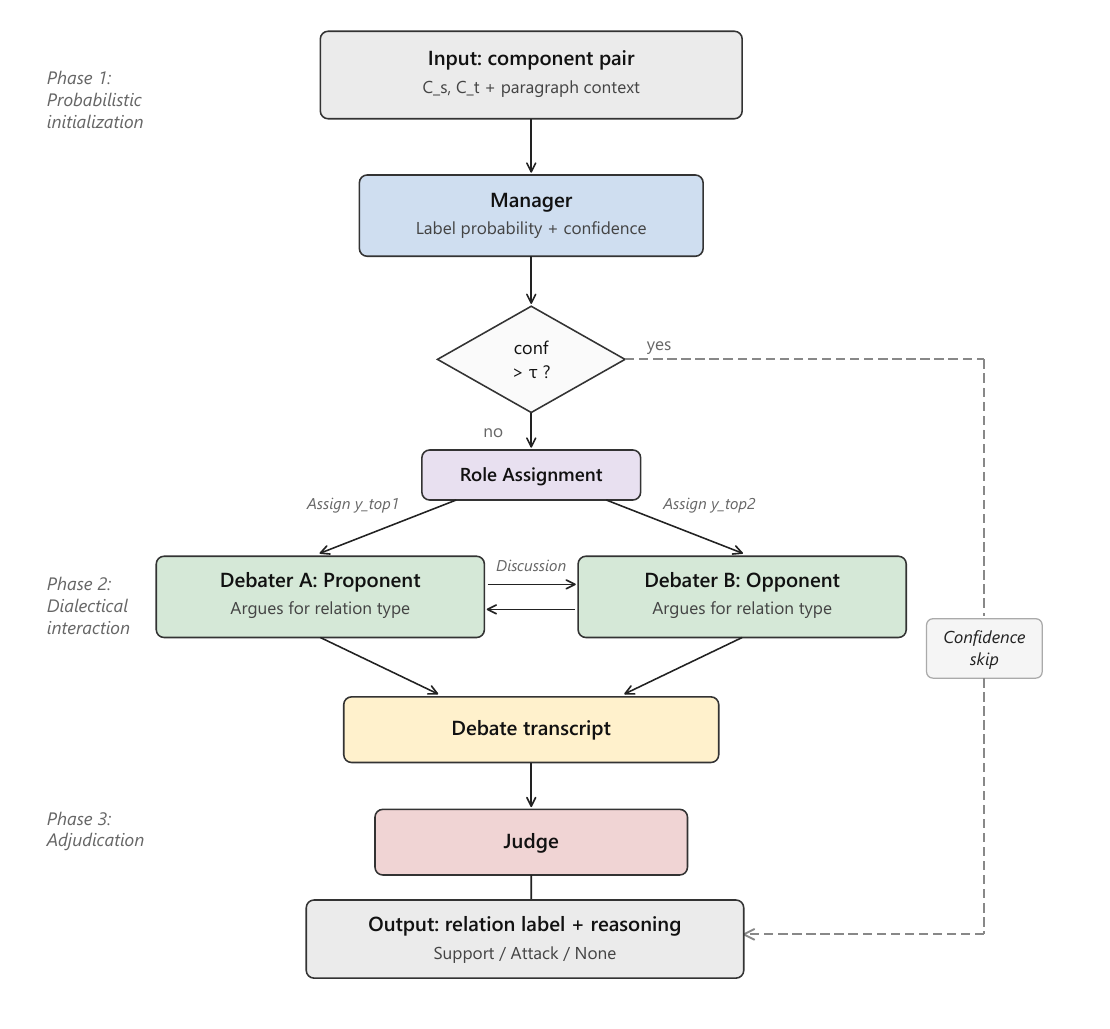}
    \caption{Overview of the proposed framework, adapted from~\cite{baba2026multiagentdialecticalrefinementenhanced} for the relation classification task. When the Manager's confidence exceeds threshold $\tau$, the debate is skipped. Otherwise, after $R$ rounds of dialectical interaction, the Judge renders a verdict based on the debate transcript.}
    \label{fig:framework}
\end{figure}

We extend the multi-agent debate architecture introduced in~\cite{baba2026multiagentdialecticalrefinementenhanced} to pair-wise relation classification. The framework retains the same agent set $\mathcal{A} = \{\text{Mgr}, \text{Prop}, \text{Opp}, \text{Jud}\}$, role-specific prompts $\mathcal{P}$, and shared transcript $\mathcal{T}$. The adaptations target the input representation (component pairs rather than single components), the label set ($\mathcal{Y}$), and the reasoning directives given to each agent. Execution proceeds in three phases, as illustrated in Figure~\ref{fig:framework}.

\paragraph{Phase 1: Probabilistic initialization}
The \textbf{Manager} estimates label probabilities for the input pair. The two most probable labels are randomly assigned to the Proponent and Opponent to mitigate position bias. If the Manager's confidence for the top label exceeds a threshold $\tau$, the debate is skipped entirely, and the label is assigned directly (\textit{confidence gating}). This mechanism, not present in the original framework, substantially reduces computational cost for unambiguous cases and is analyzed in Section~4.2.

\paragraph{Phase 2: Dialectical interaction}
The debate proceeds for $R$ rounds ($2R$ total turns). Unlike the component-level setting in~\cite{baba2026multiagentdialecticalrefinementenhanced}, where debaters argue about what a component \textit{is}, relation classification requires reasoning about what \textit{connects} two components — the direction of logical dependency, the presence or absence of evidential support, and whether an apparent link is genuinely grounded in the text. When the \textit{None} label is debated, the assigned agent must argue for the absence of any relation, providing structural evidence that no logical link exists. This adversarial dynamic is designed to expose hallucinated reasoning - where a model constructs plausible-sounding but textually ungrounded justifications for a relation.

\paragraph{Phase 3: Judge classification}
The \textbf{Judge} evaluates the full debate transcript and assigns the final label. It is specifically instructed to penalize unsupported logical jumps — claims of relations not grounded in textual evidence. The Judge defaults to \textit{None} when neither side provides sufficient grounding for a relation, directly targeting the hallucination problem.

\subsection{Experimental Setup}
\subsubsection{Dataset.} We evaluate on the \textbf{UKP Argument Annotated Essays v2}~\cite{stab2017parsing}, a corpus of 402 persuasive essays containing 6,089 argument components and their annotated relations. To ensure direct comparability with both the original framework~\cite{baba2026multiagentdialecticalrefinementenhanced} and the supervised state of the art~\cite{cabessa-etal-2025-argument}, we adopt the standard randomized test split of 80 essays established in prior literature. From the annotated essays, we extract all candidate component pairs within the same paragraph and assign each pair a label from $\mathcal{Y} = \{\textit{Support}, \textit{Attack}, \textit{None}\}$ based on the gold annotations. Pairs with no annotated relation are labeled as \textit{None}.

Prior to processing, the corpus is formatted to enable LLM agent comprehension. For each candidate pair, the full paragraph is provided with the source component marked as \texttt{<SOURCE>...</SOURCE>} and the target component as \texttt{<TARGET>...</TARGET>}.

\subsubsection{Baselines.} To validate the effectiveness of the multi-agent framework, we compare against three single-agent baselines:
\begin{enumerate}
\item \textbf{Vanilla}: Standard zero-shot prompting with the same model as the Manager agent (Gemini 2.5 Flash), using a direct classification prompt.
\item \textbf{Chain-of-Thought (CoT)}: Extends the Vanilla baseline with step-by-step reasoning prompting using Gemini 2.5 Flash, to assess whether eliciting intermediate reasoning is sufficient for relation classification.
\item \textbf{Smart Reasoning}: Uses the more capable Gemini 2.5 Pro model with built-in reasoning capabilities and the same system definitions and label descriptions provided to the Judge agent, simulating the Judge's decision-making process without the benefit of debate content.
\end{enumerate}

Additionally, we provide supervised reference baselines by fine-tuning RoBERTa-base and RoBERTa-large on the same task formulation and data split to contextualize our training-free results against standard supervised approaches.

\subsubsection{Implementation details.} The framework uses models from the Gemini 2.5 family. The \textbf{Manager} uses Gemini 2.5 Flash with temperature 0.0 to produce deterministic probability estimates over the label set. The \textbf{Debaters} use Gemini 2.5 Flash with temperature 0.7 to encourage diverse reasoning paths during argumentation. The \textbf{Judge} uses Gemini 2.5 Pro with temperature 0.0 to ensure deterministic, evidence-based adjudication. The debate length is set to 2 rounds (four total turns), allowing each debater to present their arguments and respond to counterarguments. To mitigate position bias, the two most probable labels from the Manager's distribution are randomly assigned to the Proponent and Opponent. The confidence threshold $\tau$ is varied across experiments and its effect on performance is analyzed in Section~4.2.

We evaluate performance using \textbf{Macro F1} and class-wise \textbf{Precision}, \textbf{Recall}, and \textbf{F1} scores for each relation type (Support, Attack, None).

\section{Experiments and Results}

To assess the effectiveness of the proposed framework, we design two experiments. First, we compare the confidence-gated debate against all single-agent baselines and supervised references on overall classification quality (Section~4.1). Second, we investigate the confidence gating mechanism itself, examining how the threshold $\tau$ controls the trade-off between debate usage and predictive performance (Section~4.2).

\subsection{Performance Analysis}

The central question of this experiment is whether the multi-agent debate, extended from component classification to relational reasoning, can match or outperform single-agent inference and supervised fine-tuning on the joint ARI/ARC task. Table~\ref{tab:main_results} reports the results on the UKP test set. The proposed framework is evaluated in two configurations: full debate, where all samples undergo the dialectical process, and the confidence-gated variant with $\tau = 0.70$, selected based on the analysis in Section~4.2.

\begin{table}[]
\centering
\caption{Classification performance on the UKP Student Essays test set.}
\label{tab:main_results}
\begin{tabular}{lccccc}
\toprule
\textbf{Method} & \textbf{Macro F1} & \textbf{W-F1} & \textbf{F1 (Sup.)} & \textbf{F1 (Att.)} & \textbf{F1 (None)} \\
\midrule
\multicolumn{6}{l}{\textit{Inference-only baselines}} \\
Baseline A (Vanilla)        & 0.549 & 0.658 & 0.623 & 0.340 & 0.683 \\
Baseline B (CoT)            & 0.560 & 0.684 & 0.636 & 0.327 & 0.716 \\
Baseline C (Smart Reasoning)& 0.578 & 0.666 & 0.635 & 0.412 & 0.688 \\
Proposed (full debate)      & 0.561 & 0.664 & 0.609 & 0.378 & 0.697 \\
Proposed ($\tau = 0.70$)    & \textbf{0.585} & 0.672 & \textbf{0.630} & \textbf{0.427} & \textbf{0.698} \\
\midrule
\multicolumn{6}{l}{\textit{Supervised baselines}} \\
RoBERTa-base                & 0.473 & 0.707 & 0.552 & 0.071 & 0.797 \\
RoBERTa-large               & 0.522 & 0.721 & 0.610 & 0.167 & 0.788 \\
\bottomrule
\end{tabular}
\end{table}
 
The confidence-gated variant achieves the highest Macro~F1 of 0.585 among all inference-only methods, outperforming Baseline~C (Smart Reasoning, 0.578), the strongest single-agent baseline. The improvement is concentrated on the Attack class (F1 of 0.427 vs. 0.412), the most challenging category due to its sparse representation in the corpus. However, in the test set, the Attack class contains only 42 test instances, meaning that each prediction changes this F1 score substantially. The best configuration debates only 332 out of 2{,}407 samples (14\%). In contrast, the full debate variant, which processes all samples, achieves a lower Macro~F1 of 0.561, which is lower than Baseline C.
 
The supervised RoBERTa baselines achieved the highest Weighted~F1 scores (0.707 and 0.721), reflecting strong performance on the majority None class (F1 of 0.797 and 0.788). However, even though class-weighted training was performed, both models struggled with the most underrepresented Attack pairs - 160 pairs were insufficient for reliable fine-tuning. As a result, their Macro~F1 scores fall below all inference-only approaches.

\subsection{Confidence Threshold Analysis}
\label{sec:threshold}

The confidence gating mechanism focuses on splitting samples into two groups: ones where the Manager is confident are accepted directly, while uncertain cases are the subject of a debate. This experiment examines how the threshold $\tau$ affects this split and, more importantly, classification quality. Table~\ref{tab:threshold} reports Macro~F1, Weighted~F1, Attack~F1, and the number of debated samples across threshold values.

\begin{table}[]
\centering
\caption{Effect of the confidence threshold $\tau$ on classification performance and debate usage.}
\label{tab:threshold}
\begin{tabular}{lcccc}
\toprule
\textbf{Threshold $\tau$} & \textbf{Macro F1} & \textbf{W-F1} & \textbf{F1 (Att.)} & \textbf{Debated} \\
\midrule
Manager only        & 0.562 & 0.656 & 0.383 & 0 (0\%) \\
$> 0.50$            & 0.571 & 0.664 & 0.402 & 168 (7\%) \\
$> 0.60$            & 0.583 & 0.670 & 0.425 & 286 (12\%) \\
$> 0.65$            & 0.584 & 0.669 & 0.427 & 309 (13\%) \\
$> 0.70$            & \textbf{0.585} & 0.672 & \textbf{0.427} & 332 (14\%) \\
$> 0.75$            & 0.571 & 0.680 & 0.374 & 1160 (48\%) \\
$> 0.80$            & 0.570 & 0.678 & 0.374 & 1184 (49\%) \\
Full debate         & 0.561 & 0.664 & 0.378 & 2407 (100\%) \\
\bottomrule
\end{tabular}
\end{table}

The Manager alone ($\tau = 0.0$) achieves a Macro~F1 of 0.562. As the threshold increases from to 0.70, Macro~F1 rises steadily to 0.585 while the debated proportion grows from to 14\%. At $\tau = 0.70$, the debate changed 84 predictions for the better and 48 for the worse among the 332 debated samples, yielding a net positive effect of 36 corrections.
 
However, beyond $\tau = 0.70$, a sharp transition occurs. At $\tau = 0.75$, the debated proportion jumps to 48\% while Macro~F1 drops back to 0.571 and Attack~F1 falls from 0.427 to 0.374. Full debate ($\tau = 1.0$) shows 249 improvements against 236 regressions across all 2407 samples, resulting in a near-zero net benefit. The gating at $\tau = 0.70$ reduces the total number of debate initializations by approximately 86\%.

\section{Discussion and Evaluation}
 
The experimental results confirm that the multi-agent debate approach transfers from component classification and can be used for relational reasoning, however, relational usage is conditional. In this section we interpret and evaluate the results and address the limitations of the current approach.
 
\subsection{Overall Performance}
 
All generative approaches substantially outperform RoBERTa models on Macro~F1, despite them being fine-tuned - having access to the full training split. This gap is driven almost entirely by the Attack class: RoBERTa-base achieves an Attack~F1 of just 0.071 and RoBERTa-large 0.167, while even the weakest generative baseline (Vanilla) reaches 0.340. The class-weighted loss used during fine-tuning was insufficient to compensate for 160 training instances. On the other side, generative models bring their general linguistic knowledge about what attack and contradiction look like, making them superior at handling this minority class without any task-specific examples. This confirms that the supervised approaches rely heavily on annotated data, and in domains where it is limited, training-free generative approaches offer a practical alternative. This shows direct opportunity for argument mining in low-resource languages and specialized domains, where corpora with balanced class distributions are rare. 
 
Comparing with MAD-ACC~\cite{baba2026multiagentdialecticalrefinementenhanced}, where the debate achieved a cleaner improvement on component classification (Macro~F1 of 85.7\% vs.\ 84.9\% for the strongest single-agent baseline), the margin here is not unconditional and it depends on confidence gating. This is consistent with the increased difficulty of relational reasoning: it requires simultaneous reasoning about two components and their logical dependency, rather than classifying a single span. Additionally, the None class dominates the label distribution (2/3 of test pairs), consequently creating more opportunities for the debate to predict non-existent links.
 
\subsection{Effectiveness of the Debate}
 
The most important finding of this work is that unconditional debate does not help - and can actively hurt. The full debate configuration changed 249 predictions for the better but 236 for the worse, yielding a near-zero net benefit and a Macro~F1 below Baseline~C. This aligns with the analysis of Choi et al.~\cite{choi2025debatevoteyieldsbetter}, who showed that multi-agent debate alone does not reliably improve performance, and targeted architectural interventions are needed. In our framework, just assigning specific roles to the agents wasn't enough - instead, we introduced confidence gating which precisely served this role. By restricting debate to the >70\% threshold, we reduced number of debates to only 14\% of samples where the Manager is least certain. As a result, it concentrates dialectical effort where it adds value.
 
The margin band analysis in Section~4.2 reveals the mechanism behind this effect. Below a confidence margin of 0.6, the debate produces a strong net positive effect (54 improvements vs. 41 regressions). These are cases where the Manager is highly uncertain and the adversarial interaction surfaces evidence that a single pass misses. Above a margin of 0.8, the pattern reverses: the debate changes 49 predictions, of which 62\% are for the worse. At this level, the Manager was usually already correct, and the debate introduced doubt where it was not initially present, resulting in one of the debaters constructing a superficially plausible counterargument that misleads the Judge.
 
This is a key observation for multi-agent debate systems: the value of discussion depends on whether the initial prediction is uncertain. Applying debate adds a level of uncertainty to all of the cases, even to those where it can be ignored. Not only does it legitimize the highly unlikely relation, degrading predictions, but also wastes computational resources. The confidence gate transforms the debate from a universal procedure into a tool that is used only when it is actually needed.

\subsection{Limitations and Future Work}
 
The margin of the confidence-gated framework over Baseline~C (Macro~F1 of 0.585 vs. 0.578) is small and has not been tested for statistical significance. The evaluation is limited to a single corpus (UKP Argument Annotated Essays), and as a result cross-domain generalization, while being a theoretical advantage, hasn't been empirically validated. Moreover, framework assumes gold-standard component boundaries, but in a full end-to-end pipeline errors from incorrect component detection would propagate into the relation classification stage. Finally, the debate introduces latency and cost proportional to the number of rounds and the number of pairs that require deliberation, though the confidence gate reduces this by 86\%.

These limitations are natural starting points for future work. Cross-domain evaluation on legal, political, or biomedical corpora would directly test the aforementioned generalization claim. The confidence gating mechanism itself could be a subject of a further study - by investigating the threshold impact across different datasets or replacing it with a learned gating function, the debate could be utilized even better. Finally, integrating the proposed framework with MAD-ACC into a unified end-to-end pipeline - connecting component detection with relation classification into the construction of a full graph - would result in an increased usefulness of the process, especially when deployed as a part of an automated feedback (e.g. in the educational settings).

\section{Conclusion}

We presented an extension of the multi-agent debate framework from Argument Component Classification to the Argument Relation Identification and Classification - a task where the system must reason between pairs of the components, opposed to looking at the individual spans. 

Our experiments on the UKP Argument Annotated Essays v2 corpus reveal that debate is not the definitive answer, but it can be beneficial for the relational task. Indiscriminate debate over all samples does not improve performance, as it changes nearly as many predictions for the worse as for the better, and it yields a Macro F1 score below the strongest single-agent baseline. However, when we combine debate with a confidence gating mechanism, the framework achieves the highest Macro F1 (0.585) among all training-free methods, while simultaneously reducing the number of debate initializations by 86\%. This mechanism effectively restricts debate only to the samples where the Manager is least certain. This finding aligns with recent studies showing that multi-agent debate requires targeted architectural interventions to be effective, and demonstrates that confidence-based routing can serve this role.

A secondary but notable finding is that in our experiments all generative approaches outperform fine-tuned RoBERTa models on Macro F1, despite the supervised baselines achieving higher Weighted F1. The 160 Attack training instances proved insufficient for reliable fine-tuning, while prompted models handled this minority class through general linguistic knowledge alone - reinforcing the practical value of training-free methods where annotated data is limited or heavily imbalanced. 

As a byproduct of the dialectical process, the framework produces human-readable debate transcripts that expose the reasoning behind each prediction, however, a systematic evaluation of their interpretability value remains for the future work. More broadly, by demonstrating that dialectical refinement can be conditionally effective for relational reasoning, this work takes a step toward training-free argument relation classification that is essential for applications ranging from educational writing tools to the fact checking systems.

\appendix

\bibliographystyle{elsarticle-num}
\bibliography{bibliography}

\clearpage

\end{document}